\documentclass[conference]{IEEEtran}
\IEEEoverridecommandlockouts

\usepackage[numbers,sort&compress]{natbib}
\usepackage{amsmath,amssymb,amsfonts}
\usepackage{graphicx}
\usepackage{textcomp}
\usepackage{xcolor}
\usepackage{amsmath}
\usepackage{booktabs}
\usepackage{arydshln} 
\usepackage{setspace} 
\usepackage{hhline}
\usepackage{array} 
\usepackage{setspace} 
\usepackage{graphicx}
\usepackage{amsmath}
\usepackage{romanbar}
\usepackage{mdframed}
\usepackage{verbatim}
\usepackage{amsmath}
\usepackage{bm}
\usepackage[linesnumbered,ruled,vlined]{algorithm2e}
\usepackage{cleveref}
\usepackage{mathrsfs}
\creflabelformat{equation}{#2#1#3}
\usepackage{arydshln}
\usepackage{algpseudocode}

 \usepackage{tikz}
\usetikzlibrary{shapes,arrows}

\tikzstyle{dialogue} = [
  rectangle, 
  draw, 
  text width=0.5\textwidth, 
  minimum height=2em, 
 align=left,
  fill=yellow!7
]

\def\BibTeX{{\rm B\kern-.05em{\sc i\kern-.025em b}\kern-.08em
    T\kern-.1667em\lower.7ex\hbox{E}\kern-.125emX}}

\begin{document}

\title{DiffNAS: Bootstrapping Diffusion Models by Prompting for Better Architectures\\
\thanks{*corresponding author}
}

\author{
    Wenhao Li\textsuperscript{1}, Xiu Su\textsuperscript{2*}, Shan You\textsuperscript{1}, Fei Wang\textsuperscript{1}, Chen Qian\textsuperscript{1}, Chang Xu\textsuperscript{2}\\

    \textsuperscript{1}\textit{Sensetime Research} 
    \textsuperscript{2}\textit{The University of Sydney}
}

\maketitle

\begin{abstract}
Diffusion models have recently exhibited remarkable performance on synthetic data. After a diffusion path is selected, a base model, such as UNet, operates as a denoising autoencoder, primarily predicting noises that need to be eliminated step by step. Consequently, it is crucial to employ a model that aligns with the expected budgets to facilitate superior synthetic performance. In this paper, we meticulously analyze the diffusion model and engineer a base model search approach, denoted "DiffNAS". Specifically, we leverage GPT-4 as a supernet to expedite the search, supplemented with a search memory to enhance the results. Moreover, we employ RFID as a proxy to promptly rank the experimental outcomes produced by GPT-4. We also adopt a rapid-convergence training strategy to boost search efficiency. Rigorous experimentation corroborates that our algorithm can augment the search efficiency by 2$\times$ under GPT-based scenarios, while also attaining a performance of 2.82 with 0.37 improvement in FID on CIFAR10 relative to the benchmark IDDPM algorithm.

\end{abstract}


\section{Introduction}

Image synthesis, a critical application within the field of deep learning, has seen extensive experimentation within generative adversarial networks (GANs) \cite{b1}, autoregressive models \cite{b65},  and variational autoencoders (VAEs) \cite{b67}. Lately, Diffusion Probabilistic Models (DDPM) \cite{b5}, parameterized via Markov Chain Training and reversed with a diffusion process, have demonstrated impressive outcomes in image synthesis and beyond. The state-of-the-art results and the significant variability generated by these autoencoders have sparked interest among researchers to enhance and study related frameworks. However, when the DDPM framework is fixed, the foundational model, such as Unet \cite{b33}, assumes a pivotal role in the reversed diffusion process. In this paper, we delve into the profound and consequential implications of enhancing the foundational model on the diffusion process.


Diffusion models, which are extensively used in a variety of generative tasks including image generation \cite{b50}, editing \cite{b8}, and video generation \cite{b11}, hinge on two processes: forward diffusion and reverse denoising. Models such as the text-to-image Stable Diffusion \cite{b13} have gained considerable interest due to their superior generative capabilities and user-friendly applications. In the forward diffusion process, the data distribution is progressively transformed into Gaussian noise through systematic noise additions. Conversely, the reverse denoising process, often facilitated by a base model such as Unet , approximates the inverse of the forward process, thereby enabling the conversion from Gaussian noise back to the target distribution. Given this established diffusion framework, the quality and performance of image synthesis largely hinge on the feature learning capabilities of the base model.

Given the widespread application of diffusion models, the optimization and training of these models have gradually become hot research topics. Existing strategies for diffusion model optimization include the use of learnable variance, refinement of the noise-adding strategy \cite{b6}, transfer of diffusion from pixel space to latent space, and enhancement of the variational evidence lower bound (ELBO) to approximate log-likelihood with greater confidence \cite{b14}. However, current improvements rarely focus on the optimization of the denoising autoencoder within these models. A considerable majority of existing diffusion models employ Unet as the denoising autoencoder, but these Unet models lack a rigorous design specifically tailored to diffusion models. The Unet models in current use rely primarily on empirical and heuristic manual design, significantly impeding the potential for higher generative capacities within diffusion models. To optimize the Unet architecture, an intuitive approach is to employ Neural Architecture Search (NAS) \cite{su2021prioritized,b23}, exploring an Unet-based architecture that is optimally suited for diffusion models.

The primary objective of NAS is to identify the most effective model architecture tailored to a specific task, constrained within a pre-defined search space. NAS can autonomously unearth the optimal architecture through the application of specific rules, thus eliminating the necessity for expert intervention and comprehensive manual experimentation. Presently, the majority of efficient NAS algorithms \cite{su2022vitas,su2021bcnet,cao2023re} predominantly adopt weight sharing strategies. These methods have a lower upper limit on their effectiveness and require training a super-network, which is a particularly challenging task for diffusion models.

GPT-4 \cite{b16} currently represents the pinnacle of general-purpose large language models, demonstrating exceptional proficiency in natural language processing and generation. It has been trained on a diverse corpus of general and domain-specific data, thus achieving mastery over a broad spectrum of professional tasks. In light of GPT-4's extensive training data, which encompasses profound knowledge of deep model architectures and NAS, we posit that it could supersede the supernet as a proxy for NAS.

In this paper, we undertake a comprehensive investigation of the diffusion model within the context of the UNet framework, resulting in our proposed solution, DiffNAS. Specifically, we elaborate on our strategy that combines the NAS algorithm to pinpoint the optimal foundation for the diffusion model. We employ GPT-4 as a supernet to accelerate the architecture search and maintain a search memory repository to enhance the diversity of our search outcomes. In addition, we introduce the use of Rethinking FID (RFID) as a proxy to expedite the ranking of candidate architectures, thereby circumventing the need for exhaustive training of each individual candidate. This technique is further enriched by a rapid-convergence training strategy, specifically designed to augment the efficiency of the search process. Experimental validation shows that this training strategy can achieve higher accuracy in performance ranking with only half of the training cost.

\section{Related Work}
\subsection{Improvement of Diffusion Models}
Primarily, improvements of diffusion models target towards two aspects: accelerating the sampling speed and boosting the quality of samples.

The methods for acceleration primarily fall into two categories. The first one involves refining the sampling algorithm by eliminating the Markov chain in the DDPM (Denoising Diffusion Probabilistic Models),  thus enabling fewer sampling steps than the diffusion steps used during training \cite{b18}. The second one utilizes a knowledge distillation method, in which a 'student' diffusion model learns to mimic the multiple-step denoising of a 'teacher' diffusion model in a single step, thereby allowing a diffusion model with fewer steps to achieve performance typically requiring thousands of diffusion steps.

Regarding the improvement of generation quality, the first strategy is to use learnable variance. In DDPM, the variance in the denoising process is determined by the variance in the forward diffusion process, which restricts the optimization space of the denoising autoencoder. Consequently, diffusion models such as Improved Denoising Diffusion Probabilistic Models (IDDPM) \cite{b6} employ learnable variance to enhance the diffusion model's ability to fit the target distribution. Another strategy involves employing an enhanced optimization objective. Given the difficulty of directly optimizing the actual target of diffusion model optimization—log likelihood, DDPM optimizes the variational lower bound, which has a variational gap with log likelyhood. Therefore, models like IDDPM optimize the learning objective to approximate the log likelihood more closely.
\begin{figure*}[ht]
\centering
\includegraphics[width=\textwidth]{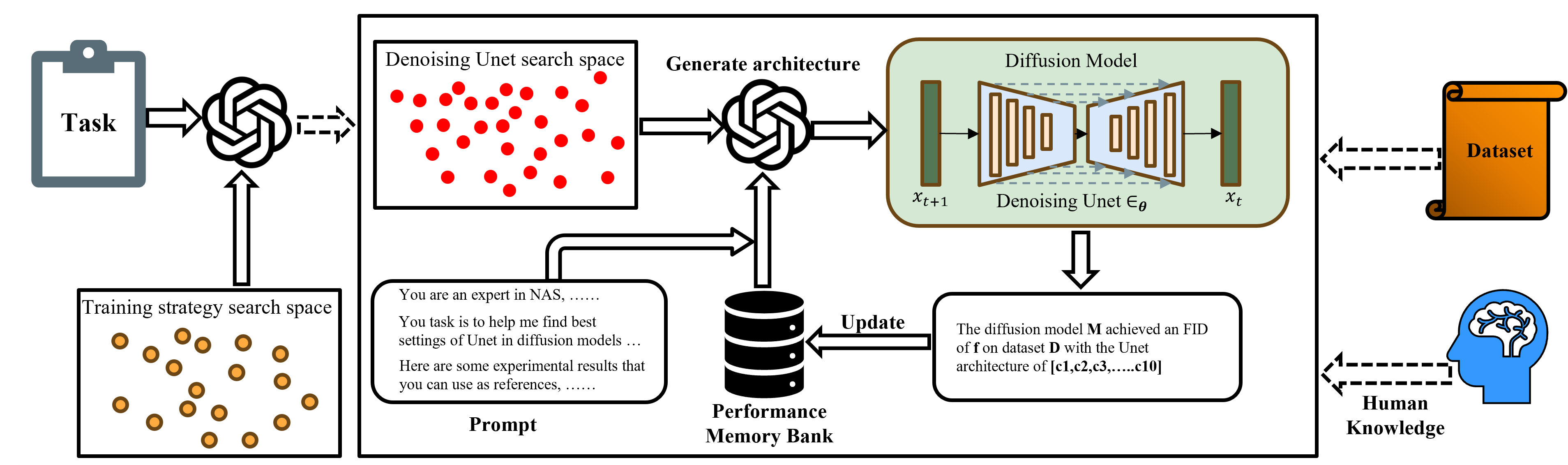}
\caption{Framework of proposed  GPT-4-Driven search for optimal Unet architecture of diffusion models}
\label{fig1}
\end{figure*}
\subsection{Neural Architecture Search}
Designing a well-performing network architecture requires professional expertise and is also very time-consuming. The purpose of NAS algorithms is to automate this process, thereby reducing the workload of humans. The workflow of NAS algorithms is to search for an architecture within predefined search space and then evaluate its performance. The performance of this architecture is used as a reference to search for a new architecture, until obtaining a network architecture that meets the requirements. 

Traditional search algorithms mainly include those based on reinforcement learning \cite{b23}, Bayesian optimization methods \cite{b24}, and gradient-based methods \cite{b26}. Compared to these algorithms, the algorithm we propose based on GPT-4 \cite{b16} is less training-intensive and faster in search speed. 

\section{revisiting diffusion model with foundation models}
Diffusion probability model is a kind of latent variable model, with the workflow composed of a forward diffusion process and a reverse denoising process. 
\subsection{Forward Process}
The original data distribution can be defined as $q(\mathbf{x}_0)$. In the forward diffusion process, given an original data $\mathbf{x}_0\sim q(\mathbf{x}_0)$, a total of $T$ diffusion steps are undertaken, getting $T$ noisy latent variables,$\mathbf{x}_1,\mathbf{x}_2,\mathbf{x}_3...\mathbf{x}_T$, each possessing the same dimension as $\mathbf{x}_0$. The diffusion step $t$ is achieved by adding Gaussian noise to $\mathbf{x}_{t-1}$:
\begin{equation}
q(\mathbf{x}_t|\mathbf{x}_{t-1}):=\mathcal{N}(\mathbf{x}_t;\sqrt{1-\beta_t}\mathbf{x}_{t-1},\beta_t\mathbf{I})\label{eq1}
\end{equation}
where $\beta_t$ is a small positive constant between 0 and 1, representing the variance of each diffusion step. 
\subsection{Reverse Denoising Process}
It can be easily observed that when $T$ is large enough, the final $\mathbf{x}_T$ completely transforms from the original data into random Gaussian noise. Consequently, in the reverse denoising process, an $\mathbf{x}_T\sim \mathcal{N}(0,1)$ is sampled from the standard Gaussian distribution as the initial noise. If a neural network can fit the reverse process of the forward diffusion, this random noise $\mathbf{x}_T$ can be transformed into a target sample. The reverse distribution from $\mathbf{x}_t$ to $\mathbf{x}_{t-1}$ is defined as follows:
\begin{equation}
p_\theta(\mathbf{x}_{t-1}|\mathbf{x}_t):=\mathcal{N}(\mathbf{x}_{t-1};\mu_\theta(\mathbf{x}_t,t),\sigma_t^2\mathbf{I})\label{eq2}
\end{equation}
where $\mu_\theta(x_t,t)$ is the mean of reverse distribution obtained from a neural network, and $\sigma_t$ is the variance of the reverse distribution obtained from $\beta_t$:
\begin{equation}
\sigma_t=\sqrt{1-(\prod_{i=1}^t\sqrt{1-\beta_i})^2}\label{eq3}
\end{equation}

\subsection{Training and Sampling}
In the practical application of diffusion models, researchers use denoising autoencoders to directly predict the noise that needs to be eliminated at each step, which yields better results than predicting the mean of distribution. In the $k$-th iteration of training, parameters of denoising autoencoder are updated as follows: 
\begin{equation}
\theta_k=\theta_{k-1}-\lambda \nabla_\theta||\mathbf{\epsilon}-\mathcal{U}_\theta(\sqrt{\overline{\alpha}_t}\mathbf{x}_0+\sqrt{1-\overline{\alpha}_t}\mathbf{\epsilon}, t)||^2 \label{eq4}
\end{equation}
where $\alpha_t = 1-\beta_t$, $\overline{\alpha}_t = \prod_{i=1}^t \alpha_i$, $\lambda$ is the learning rate, $\theta$ represents parameters of denoising autoencoder $\mathcal{U}$, $t$ is the diffusion step and $\mathbf{\epsilon}$ is a random noise sampled from Standard Gaussian Distribution $\mathcal{N}(0,1)$.

Once we have a well-trained denoising autoencoder, we can use it to gradually transform the noise $\mathbf{x}_T$ into the sample $\mathbf{x}_0$.
\begin{equation}
\mathbf{x}_{t-1}=\frac{1}{\sqrt{\alpha_t}} (\mathbf{x}_t-\frac{1-\alpha_t}{\sqrt{1-\overline{\alpha}_t}}\mathcal{U}_\theta(\mathbf{x}_t,t))+\sigma_t \mathbf{\epsilon}\label{eq5}
\end{equation}

\section{DiffNAS}
According to the information provided by  Eq. \eqref{eq2} and  Eq. \eqref{eq5}, the performance of the diffusion model primarily depends on the capacity of the foundation model $\mathcal{U}_\theta$, given other configurations of the diffusion model algorithm are held constant. Currently, almost all mainstream diffusion models such as Guided Diffusion\cite{b34} and Imagen\cite{b35} leverage Unet as their foundation model to carry out denoising in the reverse process. However, it should be noted that Unet was originally proposed for medical image segmentation tasks\cite{b33} and was not specifically designed to function as a denoising autoencoder. With this in mind, we are exploring the possibility of designing a highly effective, low-latency Unet architecture that is specifically optimized for the diffusion model's needs.
\subsection{Problem Formulation}
Given a dataset $\mathcal{D}$, diffusion model $\mathcal{M}$ is determined by a denoising autoencoder Unet $\mathcal{U}$ and other settings $\mathcal{S}$, where $\mathcal{S}$ includes elements like the diffusion steps and noise schedule. With the fixed settings $\mathcal{S}$, our task is to obtain an optimal foundation model within pre-defined search space $\mathcal{A}$. Exhausted search is as follows:
\begin{equation}
\begin{aligned}
&\mathcal{U}^* = \mathop{\arg\min}_{\mathcal{U} \in \mathcal{A}}~\mathop{FID(\mathcal{U}, \theta^*, \mathcal{D},\mathcal{S}) }\\
&s.t.\  \theta^* = \mathop{\arg\min}_{\theta}~ \mathcal{L}_{train}(\theta, \mathcal{U}, \mathcal{D},\mathcal{S}) \\
&\text{FLOPs}(\mathcal{U}) \leq B_0
\label{eq6}
\end{aligned} 
\end{equation}

where  $\theta$ represents the parameters of the denoising autoencoder $\mathcal{U}$, $\theta^*$ denotes the parameters after complete training, $B_0$ is the predefined FLOPs budget, FID is an evaluation metric for the diffusion model, and $\mathcal{U}^*$ is the searched optimal foundation model.

Given that the search space [base\_channel, num\_blocks, channel\_mult\_0, channel\_mult\_1, channel\_mult\_2, channel\_mult\_3, attn\_0, attn\_1, attn\_2, attn\_3] is vast, it is unfeasible to perform an exhaustive search. Hence, it becomes imperative for us to navigate this vast landscape with the aid of a proxy $\mathcal{N}$ to discover an optimal architecture of foundation model. One example of such proxies is the supernet employed by the weight-sharing NAS methods, such as the Single Path One Shot\cite{b36} and DARTS\cite{b37}. By training a supernet, we can rank the performance of various architectures within the search space and thereby identify the most optimal foundation model. Thus,  Eq. \eqref{eq6} could be updated as follows:

\begin{equation}
\begin{aligned}
&\mathcal{U}^* = \mathop{\arg\min}_{\mathcal{U} \in \mathcal{A}}~\mathop{FID}(\mathcal{N}^*(\mathcal{U}, \mathcal{D}, \mathcal{S})) \\
&s.t.\ \mathcal{N}^* = \mathop{\arg\min}_{\theta}~ \mathcal{L}_{train}(\mathcal{N}_\theta, \mathcal{D}, \mathcal{A},\mathcal{S}) \\
&\text{FLOPs}(\mathcal{U}) \leq B_0
\label{eq7}
\end{aligned}     
\end{equation}
where $\theta$ represents all the parameters of the supernet $\mathcal{N}$, and $\mathcal{N}^*$is the well-trained supernet.

\begin{figure}[t!]
\centering
\includegraphics[width=0.5\textwidth]{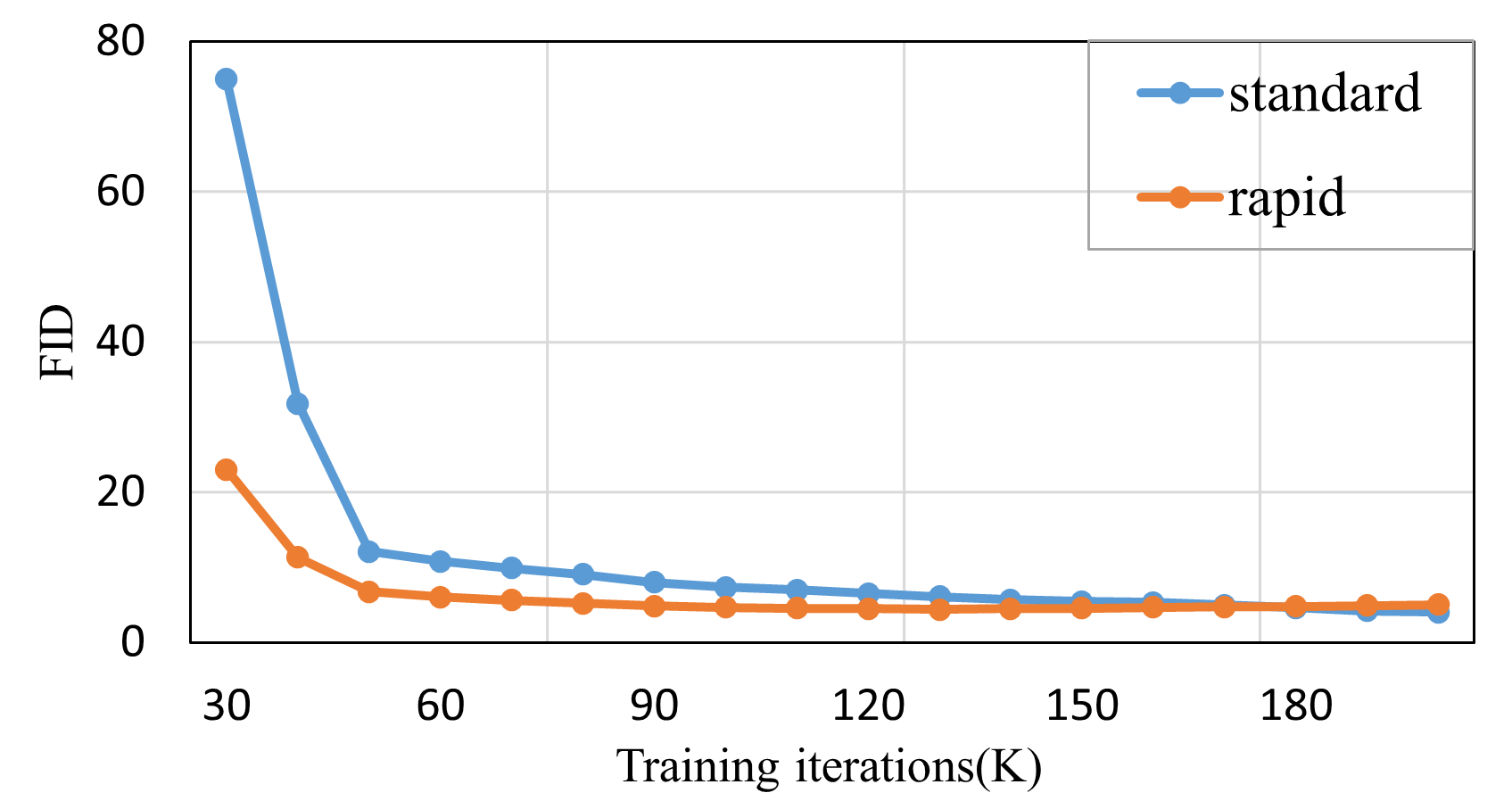}
\caption{Trend of  FID of two training strategies}
\label{fig2}
\end{figure}

However, as illustrated in Eq. \eqref{eq7}, it's necessary to train a supernet, which poses considerable challenges for diffusion models. In diffusion models, one denoising autoencoder is expected to simulate thousands of steps of reverse transformation concurrently. As it can be inferred from Fig. \ref{fig2}, the training of a diffusion model requires a significant number of iterations for convergence. During the training of a supernet on the diffusion model, employing different sub-nets at each iteration leads to the supernet's inability to converge, thereby affecting the accuracy of the ranking.

\subsection{Unet Architeture Search with GPT-4}

Given that GPT-4 has been trained on a vast quantity of data encompassing expertise from various fields, it can be directly deployed to execute numerous specialized tasks without necessitating fine-tuning. We can employ GPT as a proxy, denoted as $F_G$, for ranking the performance of foundation models. Thus, Eq. \eqref{eq6} can be updated as:
\begin{equation}
\begin{aligned}
&\mathcal{U}^* = \mathop{\arg\min}_{\mathcal{U} \in \mathcal{A}}~\mathop{FID}(F_G(\mathcal{U},\mathcal{D}, \mathcal{S})) \\
&\text{FLOPs}(\mathcal{U}) \leq B_0
\label{eq8}
\end{aligned} 
\end{equation}
However, merely using GPT-4's zero-shot ability for architecture search has significant limitations. Although GPT-4 possesses some fundamental knowledge of diffusion models and NAS, this knowledge is overly generalized for specific diffusion models and datasets, and is insufficient for precise reference. Hence, we propose an iterative search approach that conducts multiple rounds of searches. In each new round, the prior architectures, the FLOPs of the architectures, and the evaluation results are provided to GPT-4 as reference. This helps GPT-4 acquire enough prior knowledge to propose better architecture. This process can be defined as:
\begin{equation}
\begin{aligned}
&\mathcal{U}^* = \mathop{\arg\min}_{\mathcal{U} \in \mathcal{A}}~\mathop{\mbox{FID}}(F_G(\mathcal{U}, \mathcal{D}, \mathcal{S}, \mathcal{P}_i)) \\
&\text{FLOPs}(\mathcal{U}) \leq B_0
\label{eq9}
\end{aligned} 
\end{equation}
where $\mathcal{P}_i$ is a list storing the Unet architectures found in the first $i$ searches and their corresponding performance, assisting GPT-4 in understanding the effectiveness of the architectures it has proposed. The update method for this variable in the iterative search process is as follows:
\begin{equation}
\mathcal{P}_{i+1} = \mathcal{P}_i.append(\mathcal{U}_{i+1},FID(\mathcal{M}(\mathcal{U}_{i+1},\mathcal{S}),\mathcal{D}))\label{eq10}
\end{equation}
Our ultimate architecture is the one selected from $\mathcal{P}$, which exhibits the highest performance.
\begin{equation}
\mathcal{U}^* = \mathop{\arg\min}_{\mathcal{U}_i }~FID(\mathcal{U}_i), <\mathcal{U}_i, FID(\mathcal{U}_i)> \, \in \mathcal{P}\label{eq11}
\end{equation}

\subsection{\textbf{R}\textbf{a}pid-\textbf{c}onvergence \textbf{T}raining Strategy(\textbf{Ract}) }

Although the GPT-4 based method has significantly reduced computational overhead compared to traditional NAS algorithms, we still have to conduct several  searches according  to Eq. \eqref{eq8}. Each search entails a highly time-consuming training process. Our task is to discover the most effective architecture through GPT search, rather than exhaustively training a slew of decent architectures. In fact, we can use RFID (Ranking FID) to characterize the relative performance of architecture $\mathcal{U}$, while $\mathcal{Y}$ represents the training strategy, and $E$ signifies the training cost.
\begin{equation}
\begin{aligned}
&\mathcal{U}^* = \mathop{\arg\min}_{\mathcal{U} \in \mathcal{A}}~\mathop{\mbox{RFID}}(F_G(\mathcal{U}, \mathcal{D}, \mathcal{S}, \mathcal{Y}, E)) \\
&\text{FLOPs}(\mathcal{U}) \leq B_0
\label{eq12}
\end{aligned} 
\end{equation}
By improving strategy $\mathcal{Y}$ , we can achieve the objective of accurately ranking RFID with less cost $E$. Meanwhile, in Eq. \eqref{eq10}, it is not necessary to store the performance of each architecture in $\mathcal{P}_i$; instead, we store RFID. Therefore, Eq. \eqref{eq10} can be updated as follows:
\begin{equation}
\mathcal{P}_{i+1} = \mathcal{P}_i.append(\mathcal{U}_{i+1},RFID(\mathcal{M}(\mathcal{U}_{i+1},\mathcal{S}),\mathcal{D},\mathcal{Y}, E))\label{eq13}
\end{equation}
To enhance training strategy $\mathcal{Y}$ and accelerate the search process, we introduce GPT-4 Boost Augmentation to optimize the ranking accuracy of RFID. This approach enables us to achieve optimal results with the minimal training cost. To accommodate both $\mathcal{Y}$ and $E$, we can employ GPT-4 to search for $\mathcal{Y}$ based on the provision of less costly $E$, aiming to achieve the goal of rapidly generating accurate RFID. 
\begin{equation}
\begin{aligned}
&\mathcal{Y}^* = \mathop{\arg\max}_{\mathcal{Y} \in \mathcal{A}_\mathcal{Y}}~Ranking\_Accuracy(\mathcal{Y},\mathcal{D},\mathcal{S})\\
&Cost(\mathcal{Y})\leq E\label{eq14}    
\end{aligned}
\end{equation}

where $\mathcal{A}_\mathcal{Y}$ represents the search space for training strategies, encompassing searchable parameters such as learning rate, dropout, and diffusion steps. Given the relatively small search space, the search task is comparatively simple. Consequently, there is no necessity to provide GPT-4 with additional references. Merely utilizing GPT-4's zero-shot capabilities allows us to search for a training strategy that meets our requirements. The experimental results have also substantiated this assertion. Compared to standard training strategies, the training strategies discovered by GPT-4 require only half the training cost to achieve a higher ranking accuracy.

\begin{table}[t!]
\centering
\caption{performance of various models on CIFAR10}
\label{table}
\small
\setstretch{1}
\setlength{\tabcolsep}{7mm}{
\begin{tabular}{c|cc}
\toprule
model& FID & FLOPs \\
\midrule	
WaveDiff\cite{b38}& 4.01 & 6.0G \\
DiffuseVAE\cite{b40}& 3.77 &6.06G\\
DDGAN\cite{b41}& 3.75 &  5.72G\\
PNDM\cite{b43}& 3.26 &6.06G\\	
SB-FBSDE\cite{b45}& 3.01 &6.06G\\
IDDPM\cite{b6}&3.19&8.14G\\
DDPM\cite{b5}&3.17&6.06G\\ 
IDDPM+NAS&\textbf{2.82}&7.13G\\
DDPM+NAS&\textbf{2.90}&5.36G\\
\bottomrule
\end{tabular}}

\end{table}

\section{Experiments}

To verify the effectiveness of the algorithm we proposed, we conducted search experiments on two representative diffusion probabilistic models, DDPM and IDDPM. Due to limited resources, we chose to use a small dataset CIFAR10 \cite{b27} as the training dataset in the NAS experiment.

\textbf{Search Space: }The Unet used in the diffusion model has some optimizations compared to the original Unet, such as the addition of attention layers. In order to maximize the freedom of search, we have included all architecture parameters strongly related to the Unet model's capability in the search space for simultaneous search. The search space can be represented as\texttt{[base\_channel, num\_blocks, channel\_mult\_0, channel\_mult\_1, channel\_mult\_2, channel\_mult\_3, attn\_0, attn\_1, attn\_2, attn\_3]}, which has ten parameters that can be searched. $base\_channel$ represents the basic channel number of Unet, and the product of $channel\_mult\_i$ and $base\_channel$ is the channel number of the stage $i$. $num\_blocks$ represent the number of residual blocks at each stage, while $attn\_0$ to $attn\_3$ represent whether to add attention layers at each stage.

\subsection{Rapid-convergence Ablation Study}

To validate the superiority of our proposed rapid-convergence training strategy, we performed ablation experiments on IDDPM using the CIFAR10 dataset. The standard training strategy for IDDPM sets the noise schedule to cosine, with dropout configured to 0.3, learning rate set to 1e-4, diffusion steps set to 4000, batch size configured to 128, while using learnable variance. Under this configuration, a complete training of IDDPM on CIFAR10 requires a total of 0.5 million steps. We employ GPT-4 to search among the learning rate, diffusion steps, and dropout, in an attempt to discover a weak augmented training strategy to accelerate the search process. Following a single search, the training strategy proposed by GPT-4 sets the learning rate to 2e-4, dropout to 0.1, and diffusion steps to 400 steps.

\begin{table}[t!]
	\centering
	\caption{Table of ablation study}
 \label{table2}
        \setstretch{1}
\small
\setlength{\tabcolsep}{3.5mm}{
\
	\begin{tabular}{l||c|c|c}
		\toprule
		 &Spearman & Pearson & Kendall\\
		\midrule
		standard(3W) & 0.533 & 0.533 & 0.444 \\
    standard(5W) & 0.816 & 0.816 & 0.667 \\
    standard(10W) & 0.95 & 0.95 & 0.833 \\
    rapid(3W) & 0.892 & 0.892 & 0.714 \\
    rapid(5W) & 0.964 & 0.964 & 0.904 \\
		\bottomrule
	\end{tabular}}
\vspace{-3mm}
\end{table}

Our ablation study initially selected ten architectures of Unet. Subsequently, we trained ten IDDPMs each using the standard training strategy and our rapid-convergence training strategy, for 200k steps respectively. We evaluated the performance of these IDDPMs as well as their correlation with the model performance under complete training.

To evaluate the correlation between the model's performance after a small number of training steps and the performance of the fully trained model, we selected three evaluation metrics: Pearson's correlation coefficient, Spearman's rank correlation coefficient, and Kendall's tau rank correlation coefficient. The range  of values for these three metrics is from -1 to 1. A value closer to 1 indicates a stronger positive correlation, a value closer to -1 indicates a stronger negative correlation, and a value around 0 indicates no correlation.

After ten IDDPMs are fully trained using the standard training strategy, a FID performance ranking $S1$ can be obtained. Given a training strategy and the number of training steps, a performance ranking $S2$ can be obtained after the training. When three metrics between $S1$ and $S2$ are close to 1, this training strategy and number of training steps offer significant guidance for architecture search. During the calculation of the FID scores, we chose to generate 50,000 samples with 100 steps each time. 

The three metrics for different strategies and training steps are shown in TABLE \ref{table2}. From the experimental result, it can be observed that under the same training steps, our rapid-convergence training algorithm can achieve a higher correlation metrics compared to the standard training algorithm. It only requires 50k training steps to surpass the performance of 100k steps in standard training, which proves that this strategy can effectively accelerate the process of architecture search.

\subsection{Experiments on DDPM}

We first conducted search experiments on the most classic algorithm in diffusion probabilistic models, DDPM. The original DDPM uses a linear noise schedule, dropout is set to 0.1, the number of diffusion steps is 1000, and the learning rate is 1e-4. In the original configuration of Unet, the base channel is 128, the number of blocks in each stage is 2, the channel\_mult of the four stages are 1, 2, 2, 2, and the attention layer is only added in the second stage. Under this configuration, DDPM needs to be trained for 0.8M steps.

In our experiments, we first calculated the FLOPs of the original DDPM, which was 6.06G. In the search process, we constrainted the upper limit of FLOPs to 6.06G  and conducted ten rounds of searches. In order to accelerate the search process, we trained each DDPM 160k steps using rapid-convergence algorithm. 

The optimal UNet configuration identified through search has the following settings: the base channel is 96, the number of blocks in each stage is 3, the channel\_mult of the four stages are 1, 2, 3, 3, and the attention layer is added in the second and fourth stage. After fully trained, it achieves an FID of 2.90 with 5.36G FLOPs, surpassing the original architecture's FID of 3.17 with 6.06G FLOPs.

\subsection{Experiments on IDDPM}
Further, we conducted experiments on IDDPM to demonstrate that DiffNAS algorithm can be effective on a broader range of diffusion probabilistic models.

We conducted a search experiment on IDDPM using the CIFAR10 dataset. The original Unet configuration of IDDPM on CIFAR10 is: the base channel is 128, the number of blocks is 3, and the channel\_mult of the four stages are 1, 2, 2, 2, with attention layers added in the second and third stages. 

During training, we chose the cosine noise schedule, used the rapid-convergence training strategy, which need to trained for 50k steps, while original IDDPM needed 500k steps of training.  Similarly, during the ten iterative search processes, we constrainted the upper limit of FLOPs to the FLOPs of the original Unet: 8.14.

The best UNet configuration discovered through search has the following settings: the base channel is 96, the number of blocks in each stage is 4, the channel\_mult of the four stages are 1, 2, 3, 3, and the attention layer is added in the first, second and third stage. It achieves an FID of 2.82 under 7.13G FLOPs, which represent significant improvements compared to the original IDDPM's FID of 3.19 under 8.137G FLOPs.

\section*{Conclusion }
In this paper, we propose DiffNAS, a novel architecture search framework for diffusion models. Using GPT-4 as a proxy, DiffNAS efficiently discovers optimal Unet architectures through just a few dialogue rounds. With our Ract training strategy, we accelerate the search process by a factor of two and achieve superior performance. Our approach improves the FID scores on CIFAR-10 for IDDPM and DDPM by 0.37 and 0.27 respectively. In summary, our algorithm can significantly enhance the generative performance of diffusion models. Additionally, it is compatible with other optimization strategies aimed at enhancing diffusion models, showcasing its high practical value.

\small
\bibliographystyle{IEEEtran}
\bibliography{myref}

\vspace{12pt}

\end{document}